# Digraphie des langues ouest africaines : Latin2Ajami : un algorithme de translittération automatique

El hadji M. Fall[1], El hadji M. NGUER[1], BAO Diop Sokhna [1], Mouhamadou KHOULE[1],
Mathieu MANGEOT[2], Mame T. CISSE[3]
(1) Université Gaston Berger, BP 234 Saint Louis, Sénégal
(2) LIG, Université de Grenoble Alpes, 38400 Saint Martin D'HERES, France.
(3) ARCIV, Université Cheikh Anta Diop de Dakar, BP 5005 Dakar-Fann, Sénégal
`leuz8f@gmail.com,emnguer@ugb.edu.sn, baosokhna@hotmail.com`
`khoule.mouhamadou@ugb.edu.sn,mathieu.mangeot@imag.fr,thiernoc@gmail.com`

## Résumé

Les langues nationales du Sénégal, comme celles des pays d'Afrique de l'ouest en général, sont écrites avec deux alphabets : l'alphabet latin qui tire sa force des décrets officiels, et l'alphabet arabe complété (Ajami), très répandu et bien intégré, qui ne bénéficie que peu de soutien institutionnel.

Cette digraphie a créé deux mondes s'ignorant mutuellement. En effet, l'alphabet Ajami est généralement utilisé par les populations issues des écoles coraniques, alors que l'alphabet latin est utilisé par les populations issues de l'école publique.

Pour résoudre ce problème, il s'avère utile de mettre en place des outils de translittération entre ces deux écritures. Un travail préliminaire (Nguer, Bao-Diop, Fall, khoule, 2015) avait été effectué pour situer les problématiques, les défis et les perspectives. Ce présent travail, qui en fait suite, rentre dans ce cadre et a pour objectif l'étude et la mise en place d'un algorithme de translittération du latin vers l'Ajami.

## Abstract

The national languages of Senegal, like those of West Africa country in general, are written with two alphabets : the Latin alphabet that draws its strength from official decreesm and the completed Arabic script (Ajami), widespread and well integrated, that has little institutional support. This digraph created two worlds ignoring each other. Indeed, Ajami writing is generally used daily by populations from Koranic schools, while writing with the Latin alphabet is used by people from the public school.

To solve this problem, it is useful to establish transliteration tools between these two scriptures. Preliminary work (Nguer, Bao-Diop, Fall, khoule, 2015) was performed to locate the problems, challenges and prospects. This present work, making it subsequently fell into this. Its objective is the study and creation of a transliteration algorithm from latin towards Ajami.

MOTS-CLES : TALN, translittération, langues africaines, alphabet Ajami, alphabet latin.
KEYWORDS : NLP, transliteration, african languages, Ajami alphabet, latin alphabet.

# 1   Introduction

D'après le rapport 2014[1] de l'Organisation Internationale de la Francophonie (OIF), le nombre de francophones (réels et occasionnels) au Sénégal en 2013 est de 4 277 000 sur une population de 14 133 280. Ainsi 9 856 280 sénégalais ne comprennent pas la langue officielle (le français) pour s'informer et pour accéder aux savoirs, aux formations, aux textes juridiques, etc.,

Cela constitue un frein pour un développement économique réel et durable. Ainsi le français ne permet un accès à l'information, à la formation et à la communication qu'à une petite frange de la population. L'utilisation des langues nationales comme le wolof dans l'éducation et la formation demeure alors la seule alternative.

En effet le wolof est une langue véhiculaire parlée par plus de 80% de la population. Cette langue a longtemps été écrite en caractères arabes complétés non harmonisés (communément appelé wolofal), depuis les premiers contacts entre la population locale et la culture arabo-musulmane qui remontent au VIIIe et au IXe siècle de notre ère (Cissé M., 2006). Aujourd'hui elle est officiellement écrite avec les caractères latins depuis 1971[2] mais aussi avec les caractères coraniques harmonisés (CCH) depuis 2007. La première écriture tire sa force des décrets officiels et la seconde, bien que très répandue et bien intégrée, ne bénéficie que de peu de soutien institutionnel. Cette dernière est généralement utilisée par les populations issues des écoles coraniques dans la communication, les affaires au quotidien, la littérature (les textes religieux, la poésie, etc.), la religion (Coran et Bible en Ajami), la médecine traditionnelle religieuse, etc.

L'écriture avec les caractères latins est utilisée pour la localisation des TIC (Web, dictionnaires, outils de Windows et de Google traduits en wolof, etc.), la traduction des textes juridiques (code du commerce, constitution traduits en wolof) et religieux (Coran et Bible en wolof), l'édition, etc.

Mais derrière cette digraphie, coexistent deux mondes qui s'ignorent mutuellement. En effet, les personnes non instruites à l'un ou l'autre système n'ont pas accès aux documents existants dans le système qu'elles ignorent ou qu'elles ne maîtrisent pas.

Pour contribuer à résoudre ce problème et permettre un accès général aux connaissances (TIC, textes juridiques, religieux, etc.) par ces deux groupes de populations et ceci indépendamment du type d'écriture, il s'avère utile de mettre en place des outils automatiques pour translittérer les documents textes, les pages web, les SMS, les emails, etc. Ceci permettra, par exemple en un clic, d'avoir automatiquement en wolof Ajami un document écrit en wolof latin et vice-versa. Qui plus est, il permettra de publier dans les deux écritures, tout document écrit avec l'une des écritures.

C'est dans ce sillage que rentre ce présent article dont l'objectif principal est d'étudier et de mettre en place un algorithme de translittération automatiquement du latin vers l'Ajami.

Notre travail se fera comme suit. D'abord nous parlerons de la notion de translittération. Ensuite nous présenterons l'algorithme de translittération, puis nous montrerons son application à la translittération de document texte à travers une macro Word. Enfin nous verrons en perspectives comment utiliser cet algorithme pour la translittération de page Web, d'email et de SMS, etc.

---

[1] www.francophonie.org/IMG/pdf/repartition_des_francophones_dans_le_monde_en_2014.pdf

[2] Décret n° 2005-992 du 21 octobre 2005, relatif à l'orthographe et la séparation des mots en wolof. http://www.jo.gouv.sn/spip.php?article4802

## 2 La notion de translittération

### 2.1 Qu'est-ce que la translittération ?

La translittération (Nguer, Bao-Diop, Fall, khoule, 2015) est l'opération qui consiste à substituer à chaque graphème d'un système d'écriture un graphème ou un groupe de graphèmes d'un autre système, indépendamment de la prononciation. Un graphème est la plus petite unité contrastive dans le système d'écriture d'une langue[3]. L'opération est réversible et dépend donc du système d'écriture cible, mais pas de la langue. La Figure 1 représente un extrait du tableau de correspondance entre les alphabets des langues nationales du Sénégal.

| ARABE | WOLOF | PULAAR | SONINKE | MANDINKA | SEEREER | JOOLA | BALANT | SAAFI | Fotta-Guinea | Soso-Guinea | LATIN | Latin codepoint | SenAjami2OT | F.M.I | Original code point |
|---|---|---|---|---|---|---|---|---|---|---|---|---|---|---|---|
| X | X | X | X | X | X | X | X | X | X | X | Alif | U+0627 | ا | ا ا ا | U+0627 |
| X | X | X | X | X | X | X | X | X | X |   | b | U+0062 U+0042 | ب | ببب | U+0628 |
|   | X | X | X | X | X | X |   |   | X |   | p | U+0070 U+0050 | ݒ | ببب | U+0752 |
|   | X |   |   | X |   | X | X |   |   |   | ɓ | U+0253 U+0181 | ب | ببب | U+E000 |

Figure 1: Correspondance de quelques graphèmes des langues nationales du Sénégal

Comme le montre la Figure 2 , l'emploi de signes diacritiques permet de résoudre le problème du nombre différent de caractères entre les alphabets des deux systèmes d'écriture.

| Lettres Wolof | Valeur Unicode | Exemple en Wolof | Signification en français | Lettres ajami | Valeur Unicode | Exemple en Ajami |
|---|---|---|---|---|---|---|
| a | U+0061 | Am | avoir | اَ | U+064E | اَمْ |
| à | U+00E0 | Jàng | apprendre | اُ | U+E004 | جُنݣْ |
| c | U+0063 | car | branche | ݖ | U+0756 | نَݖْ |
| é | U+00E9 | wér | Être guéri | اِ | U+E008 | وِرْ |
| ë | U+00CB | Bët | oeil | اَ | U+E00A | نَتْ |

Figure 2: Correspondance de quelques graphèmes du wolof

---

[3] Crystal, David, 1997, The Cambridge Encyclopedia of Linguistics, second edition, UK: Cambridge University Press.

Cette opération de conversion a pour objectif premier de permettre la reconstitution automatique et univoque de l'écriture originale (que l'on appelle aussi rétro-conversion). En un mot, la translittération d'un texte translittéré doit retourner le texte original. On utilise pour cela des standards de normalisation. En guise d'exemple nous pouvons citer ISO 9:1995[4] qui est un standard international de translittération des caractères cyrilliques en caractères latins. En guise d'exemple, la Figure 3 représente la translittération d'un texte wolof latin en wolof Ajami.

```
Saar 1 : UBBIKU GA (SAAR WU NJËKK WI)
7 laaya – Laata Gàddaay ga
1. Ci turu Yàlla, miy Yërëmaakoon, di Jaglewaakoon, laay tàmbalee
2. Xeeti cant yépp ñeel na Yàlla, miy Boroom àddina si.
```

سَارْ 1 : اَبِّکُ گَ (سَارْ وُ نجَکّ وِ)

7 لأيَ – لَاتَ گَدَّايْ گَ

.1 تِ تَرُ يَلَّ، مِيْ يَرَمَاکُونْ، دِ جَگْلِوَاکُونْ، لأيْ تَمبَلِي

.2 خِيتِ نَنتْ يِپّ نِيلْ نَ يَلَّ، مِيْ بْرُومْ اَدِّنْ سِ.

Figure 3 : Exemple de texte wolof latin translittéré en wolof Ajami.

Il faut noter que la translittération n'est pas une traduction. En effet durant le processus de translittération, un mot écrit dans un système d'écriture (alphabétique ou syllabaire[5]) est transposé dans un autre, comme par exemple de l'alphabet wolof latin à l'alphabet wolof Ajami. En d'autres termes, aucune traduction n'est impliquée dans ce processus. Si le mot source ne signifie rien dans la langue en question, sa translittération n'en signifiera pas plus, même si elle pourrait donner l'impression d'être un mot dans cette langue.

## 2.2 À quoi sert la translittération ?

La translittération est utilisée dans plusieurs domaines. On peut citer l'exemple des bibliothèques où elle est utilisée, quand un utilisateur effectue une recherche ou indexe des contenus, pour retrouver l'information écrite dans un alphabet différent et la retourner dans le système d'écriture de l'utilisateur. La translittération permet aussi d'utiliser un clavier dans un alphabet pour taper un texte dans un autre alphabet. Par exemple ; il est possible grâce à cette technique d'utiliser un clavier Azerty pour taper du texte en Ajami.

Dans notre contexte, la translittération de page web permettra aux populations de faire des recherches sur internet avec l'écriture de leur choix (Ajami ou latin) et d'avoir les résultats présentés en latin ou Ajami. Quant à la translittération de document texte, elle permettra dans l'édition, à partir d'un document écrit dans un alphabet, de produire automatiquement le même document dans l'autre alphabet. Ce qui facilitera la publication dans les deux alphabets, et de là, permettre l'accès aux connaissances indépendamment de l'écriture. Enfin la translittération des

---

[4] http://www.iso.org/iso/fr/home/store/catalogue_tc/catalogue_detail.htm?csnumber=3589
[5] http://www.larousse.fr/dictionnaires/francais/syllabaire/76022?q=syllabaire#75151

emails et des SMS permettra de faciliter considérablement la communication entre les deux communautés. Mais faudrait-il disposer d'un algorithme efficient de translittération.

A notre connaissance, aucun algorithme de translittération automatique des langues du Sénégal, n'a encore été publié. Néanmoins quelques programmes de translittération ont été implémentés ça et là. Il s'agit de la macro Ajami63[6] de translittération de document texte sous Word et du programme de translittération0 écrit en Java (Gueye S. T., Fall T. G, 2011). Nous allons partir de ces derniers pour mettre en place notre algorithme de translittération. Cet algorithme appelé Latin2Ajami pourra être utilisé pour la translittération de document texte, de page web, d'email, de Sms, etc.

# 3  L'Algorithme de translittération Latin2Ajami

## 3.1  Présentation

Le principe de l'algorithme est simple, il se base sur un tableau de correspondance appelé **Glyph** dont les données proviennent d'un fichier externe modifiable. Ce tableau consiste uniquement en points d'Unicode, sous forme hexadécimale, un code romain à gauche et le code du caractère Ajami correspondant à droite, entre guillemets, et séparés par une virgule. Par exemple : "2C", "60C" correspondant aux caractères "," et "،". C.à.d. la virgule du latin et la virgule de l'arabe. Initialement le tableau **Glyph** contient 767 (&H2FF) points Unicode en romain, la macro lit le tableau de caractères ligne par ligne et change chaque valeur de **Glyph** par son équivalence en Ajami. Ce qui permet de transcrire chaque caractère romain par son équivalence en Ajami en point d'Unicode.

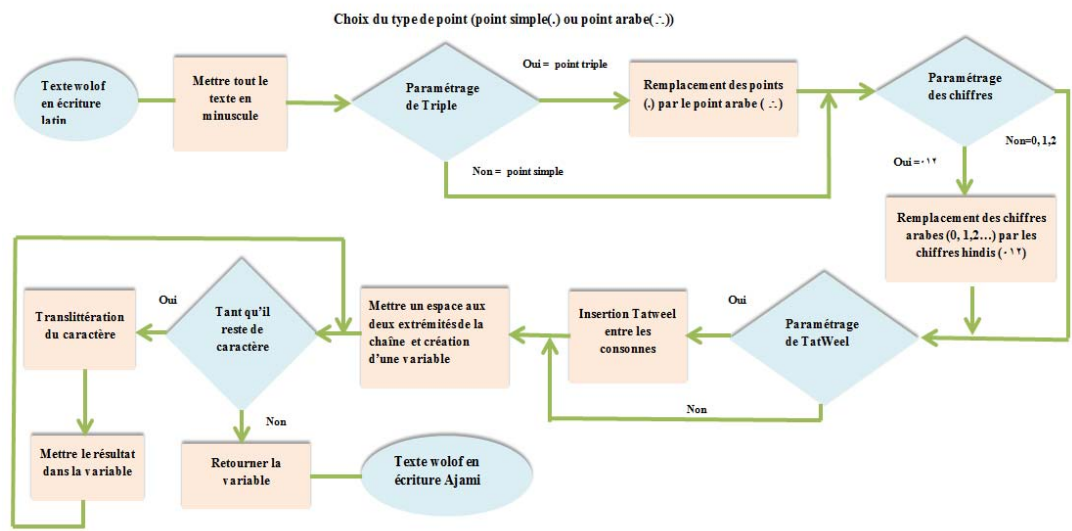

Figure 4: Organigramme général de l'algorithme Latin2Ajami

---

[6] Ajami63. http://paul-timothy.net/pages/ajamisenegal

Comme le montre l'organigramme ci-dessus, l'algorithme prend en entrée un texte latin et commence par la première opération consistant à mettre tout le texte en minuscule car l'arabe n'a pas de majuscule. Ensuite il passe par le paramétrage du point final : triple (∴) et simple (.). Notons que le point final triple se voit surtout dans les livres saints (Coran ou Bible). Si le point final triple est choisi, il remplace tous les points simples par le point triple et passe par le paramétrage des chiffres ; Sinon il passe directement par le paramétrage des chiffres : arabes (0123456789) ou Hindis (٠١٢٣٤٥٦٧٨٩). Ensuite il passe par le paramétrage de Tatweel qui sert à allonger le trait entre les consonnes. Si la réponse est oui il insère le Tatweel entre les consonnes et continue sur la création de la variable devant contenir le résultat de la translittération, sinon il passe directement à la création de la variable. A ce stade, l'algorithme rentre dans une boucle qui consiste à translittérer caractère par caractère tant qu'il en reste et met le résultat dans la variable, sinon il retourne le résultat. Cette boucle représente la fonction de translittération proprement dite.

### 3.2 La fonction de translittération proprement dite

Afin de faciliter la compréhension de la translittération proprement dite, nous avons séparé l'organigramme de la translittération des consonnes et celui des voyelles de l'organigramme général de l'algorithme.

#### *3.2.1 Translittération des consonnes*

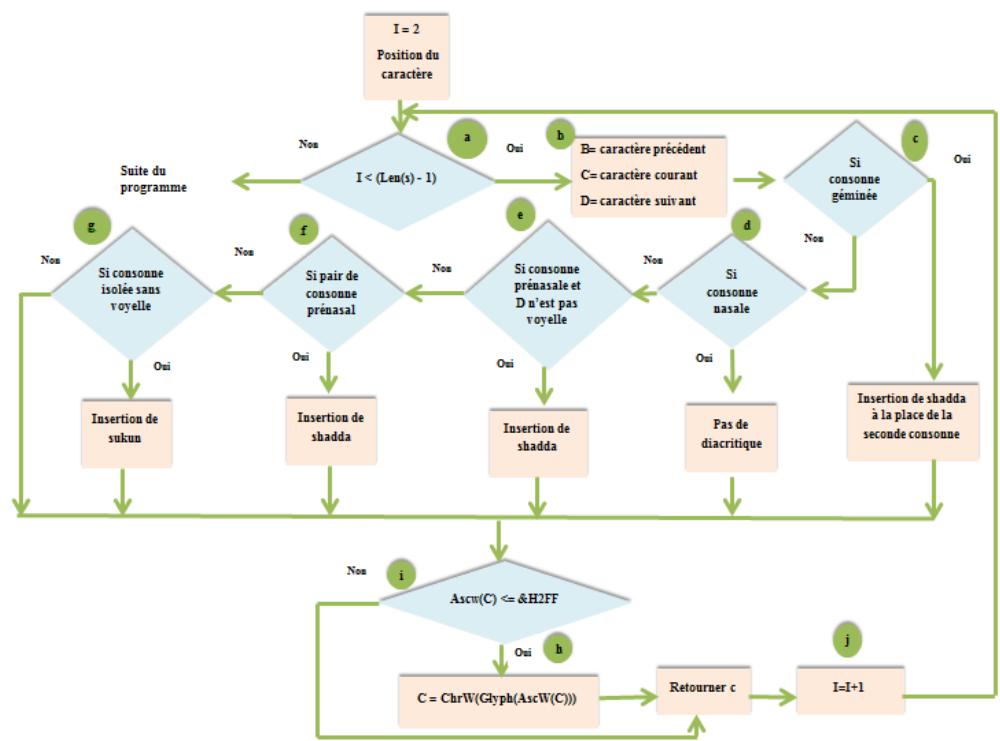

Figure 5: Organigramme de la translittération des consonnes

Comme illustré dans la Figure 5, le compteur I représentant la position du caractère courant est initialisé à 2, et les instructions suivantes sont successivement exécutées :

a) tant que I < à Len(s)-1 c.à.d. la taille de la chaîne de caractère à translittérer moins 1,
b) on affecte aux variables B, C, D respectivement au caractère précèdent, au caractère en cours et au caractère suivant.
c) Ensuite si le caractère en cours c'est-à-dire la variable C est une consonne géminée, on insert un Shadda ( ܆ ) à la place de la seconde consonne.
d) Sinon si la consonne est nasale, on ne met pas de diacritique ;
e) Sinon si elle est prénasale et le caractère D n'est pas une voyelle, on insert un Shadda ;
f) Sinon si la consonne est une paire de consonne prénasale, on insert un Shadda ;
g) Sinon si la consonne est isolée sans voyelle, on insert un sukun ( ܆ ) ;
h) Sinon si le code ASCII de la consonne est compris entre 0 et &H2FF[7], on affecte à C le code Unicode du caractère se trouvant à la position du code Ascii de C du tableau de correspondance **Glyph,** puis on retourne le caractère C.
i) Sinon on retourne directement le caractère.
j) En fin on incrémente le compteur et on répète les instructions de a) à j) jusqu'à ce que la condition de départ soit fausse.

### 3.2.2   *Translittération des voyelles*

Comme illustré dans la Figure 6 les voyelles sont traitées comme suit :

a) Si la variable C est une ponctuation ou un espace et D est une voyelle, on insert un Alif ;
b) Sinon si on a un mot commençant par aa, aa est remplacé par un madda  ( آ );
c) Sinon si on a un lam non géminé suivi d'un aa, on insert Alif avant a ;
d) Sinon si on a un lam géminé suivi par aa, on insert Alif avant a ;
e) Sinon si on a une voyelle longue, on insert une consonne à la place de la seconde voyelle.
f) Sinon si le code ASCII du caractère en cours est compris entre 0 et &H2FF, on affecte à C le code Unicode du caractère se trouvant à la position du code Ascii de C dans le tableau **Glyph**, puis on retourne le caractère C ;
g) sinon on retourne directement le caractère C.
h) Arrivé à ce stade, on incrémente le compteur et on répète les instructions de a) à h) jusqu'à ce que la condition de départ soit fausse.

---

[7]   Tous les caractères alphabétiques et numériques des langues ouest-européens ont un point d'Unicode unique compris entre 0 et &H2FF. Par exemple A=65 et Z=90.

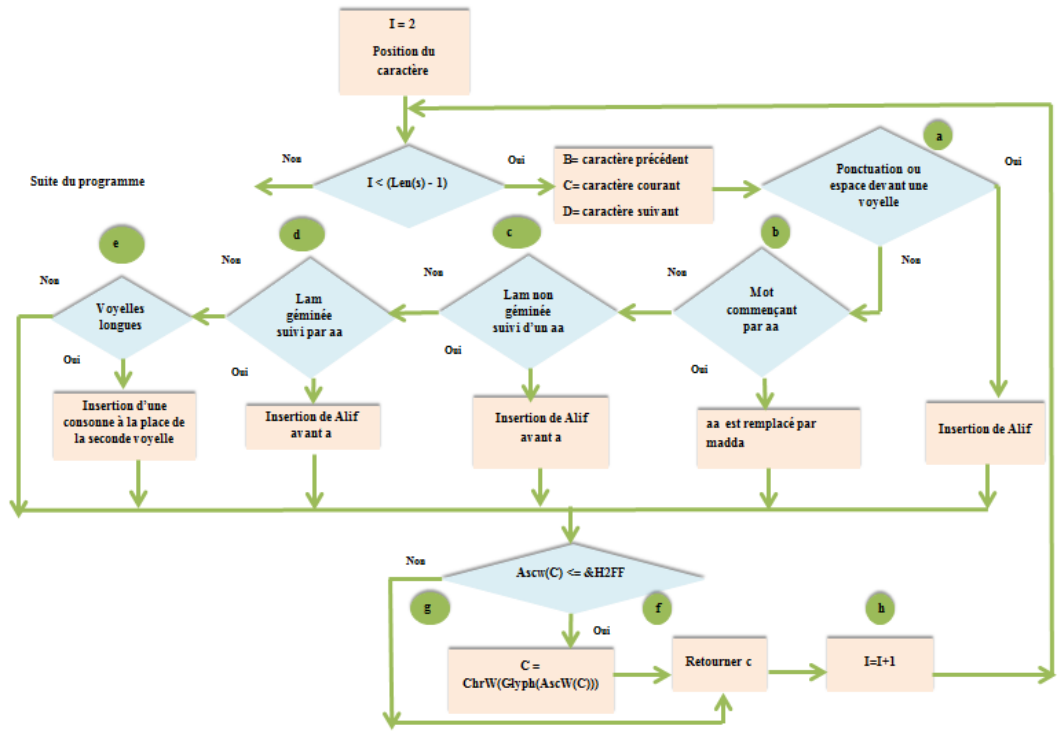

Figure 6: Organigramme de la translittération des voyelles

## 4 Application de l'algorithme Latin2Ajami

### 4.1 Translittération de document texte

L'algorithme a été appliqué pour la mise en place de la macro Ajami70 permettant la translittération de document texte sous Word. Précisons que la macro Ajami70 est conçue avec VBA (Visual Basic for Application), qu'elle translittère uniquement du texte latin en texte Ajami et qu'elle ne prend en charge que les langues wolof et sérère. Le sens inverse et les autres langues sont pris en charge dans une future proche.

La macro Ajami70 a apporté plusieurs améliorations par rapport à la macro Ajami63 en réduisant de manière conséquente le temps de translittération (100 pages en 2 minutes) au lieu de 1heure pour Ajami63 et en prenant en charge les notes de bas de page, les en-têtes et pieds de page.

**Exemple d'exécution**

L'application de l'algorithme sous VBA pour la translittération du mot *dëkkandoo* donne :

S = dëkkandoo

S = chr(32)&chr(32)&dëkkandoo&chr(32)

Len(S) =14 ;  Len(S)-1= 13

chr(32) désigne un espace en VBA.

| I | B | C | D | S | Sortie |
|---|---|---|---|---|---|
| 2 | chr(32) | chr(32) | D | | h=>chr(32) |
| 3 | chr(32) | D | Ë | | h=>د |
| 4 | D | Ë | K | | f=> ̈ |
| 5 | Ë | K | K | chr(32)&chr(32)&dëk&Shadda&andoo&chr(32) | c,h => ک |
| 6 | K | Shadda | A | | i=> ّ |
| 7 | Shadda | A | N | | f=> ́ |
| 8 | A | N | D | | d,h=> ن |
| 9 | N | D | O | | h=>د |
| 10 | D | O | O | chr(32)&chr(32)&dëk&Shadda&ando&"waw"&chr(32) | e,f=> ̊ |
| 11 | O | Waw | chr(32) | | h=>و |
| 12 | Waw | chr(32) | | | h=>chr(32) |

On obtient après reconstitution : دُكَّنْدْو

# 5   Conclusion et perspectives

Au Sénégal comme dans la plupart des pays d'Afrique de l'Ouest, la digraphie des langues locales a créé, au sein des populations parlant une même langue, deux mondes qui s'ignorent mutuellement. L'un utilise les caractères latins et l'autre utilise les caractères arabes complétés. Cette situation pose plusieurs problèmes dont une communication écrite très réduite entre ces deux mondes et l'accès limité à la connaissance et aux outils modernes de communication. C'est pour contribuer à résoudre ce problème, par la translittération automatique des deux écritures, que ce travail a été réalisé.

Ainsi, nous avons commencé le travail par présenter la notion de translittération, ensuite nous avons décrit l'algorithme Latin2Ajami de translittération du latin vers l'Ajami que nous avons mis en place. Par la suite, nous avons appliqué l'algorithme pour la translittération de document texte sous Word à travers la macro Ajami70. Cette macro qui prend en charge les notes de bas de page, les en-têtes et les pieds de page dans un document Word, permet de translittérer un document de 100 pages en deux minutes[8]. Ainsi elle pourra permettre aux chercheurs, aux écrivains et aux imprimeurs de produire très rapidement, à partir d'un document écrit en caractères latins, une version équivalente en caractères Ajami. Aussi il serait intéressant d'ajouter la prise en charge des couleurs, des images et des tableaux.

En perspectives, l'algorithme pourra être utilisé pour la translittération automatique de pages web, de Sms et d'emails afin de faciliter l'accès aux connaissances à travers le web et la communication entre les deux communautés, qui pour le moment reste principalement orale. Enfin pour permettre la communication dans les deux sens, un travail similaire devrait être effectué pour la mise en place d'un algorithme de translittération de l'Ajami vers le latin.

## Remerciements

## Références

CISSE M. (2006). Ecrits et écriture en Afrique de l'Ouest, *Revue électronique internationale de sciences du langage Sudlangues, 6* : www.sudlangues.sn/

NGUER E. M., BAO-DIOP S., FALL Y. A., KHOULE M. (2015). Digraph of Senegal's local languages: issues, challenges and prospects of their transliteration. *LTC 2015*.

NGUER E. M., KHOULE M, THIAM M. N., MBAYE B. T., THIARE O., CISSE M. T., MANGEOT M Dictionnaires wolof en ligne : état de l'art et perspectives. *CNRIA 2014*.

CHEIKHOUNA L. NGABOU, EL HADJ M. FALL (2010), *Les Caractères Coraniques Harmonisées* 2e édition 2010.

DONALD Z. OSBORN. (2005), Les langues africaines et la technologie de l'information et de communication : localiser le futur ? *27eme Conférence d'Internationalisation et d'Unicode*. Berlin, Allemagne.

GUEYE S. T., FALL T. G (2011). *La translittération automatique wolof wolofal en java*. Mémoire de maîtrise informatique Université Gaston Berger, Sénégal.

ABBAS MALIK M. G. (2010), Methods and Tools for Weak Problems of Translation.
Computer Science [cs]. Université Joseph-Fourier - Grenoble I, 2010. English. <tel-00502192>

---

[8] Sous Windows 8 sur une machine avec un processeur de 1,33GHZ et une Ram de 2Go.